\pdfoutput=1
 
\documentclass[a4paper,twoside]{article}

\usepackage{epsfig}
\usepackage{subfigure}
\usepackage{calc}
\usepackage{amssymb}
\usepackage{amstext}
\usepackage{amsmath}
\usepackage{amsthm}
\usepackage{multicol}
\usepackage{pslatex}
\usepackage{apalike}
\usepackage{glossaries}
\usepackage{color}
\usepackage{SCITEPRESS}

\newacronym{lamina}{LAMINA}{Longest set of non-overlApping Maximal correlated 
INtervAls}
\newacronym{AU}{AU}{Action Unit}
\newacronym{AUS}{AUs}{Action Units}
\newacronym{FACS}{FACS}{Facial Action Coding System}
\newacronym{S}{S}{Sender}
\newacronym{R}{R}{Receiver}
\newacronym{iou}{IoU}{Intersect over Union}
\newacronym{GC}{GC}{Granger causality}
\newacronym{BIC}{BIC}{Bayesian Information Criterion}
\newacronym{VAR}{VAR}{vector autoregressive model}

\begin{document}

\title{Causal Inference in Nonverbal Dyadic Communication with Relevant Interval 
Selection and Granger Causality}

\author{\authorname{Lea M\"uller\sup{1}, Maha Shadaydeh\sup{1}$^*$, Martin 
Th\"ummel\sup{1}, Thomas Kessler\sup{2}, Dana Schneider\sup{2} and Joachim 
Denzler\sup{1,3}}
\affiliation{\sup{1}Computer Vision Group, Friedrich Schiller University of Jena, 
Ernst-Abbe-Platz 2, 07743 Jena, Germany}
\affiliation{\sup{2}Department of Social Psychology, Friedrich Schiller University of 
Jena, Humboldtstrasse 26, 07743 Jena, Germany}
\affiliation{\sup{3}Michael Stifel Center, Ernst-Abbe-Platz 2, 07743 Jena, Germany}
\affiliation{$^*$ \email{maha.shadaydeh@uni-jena.de}}
}

\keywords{Nonverbal emotional communication, Granger causality, maximally coherent 
intervals}

\abstract{Human nonverbal emotional communication in dyadic dialogs is a process of 
mutual influence and adaptation. Identifying the direction of influence, or 
cause-effect relation between participants,  is a challenging task due to  two main 
obstacles. First,  distinct emotions might not be clearly visible. Second, 
participants cause-effect 
relation is transient and variant over time.  In this paper, we address these 
difficulties by using facial expressions that can be present even when strong distinct 
facial emotions are not visible. We also propose to apply a relevant interval selection 
approach prior to causal inference to identify those transient intervals where adaptation 
process occurs. To identify the direction of influence,  we apply the concept of Granger  
causality to the time series of facial  expressions on the set of relevant intervals. We 
tested our approach on synthetic data and then applied it to newly, experimentally 
obtained data. Here, we were able to show that a more sensitive facial expression 
detection algorithm and a relevant interval detection approach is most promising to 
reveal 
the cause-effect pattern for dyadic communication in various instructed interaction 
conditions.}

\onecolumn \maketitle \normalsize \vfill

\section{\uppercase{Introduction}}
\label{sec:introduction}

\noindent Human nonverbal communication in effective dialogs is mutual, and thus, it 
should be a process of continual two-sided adaptation and mutual influence. However, some 
humans behave consistently over time either by resisting adaptation and influence on 
purpose, or by maintaining their own style because of absent social communication 
skills \cite{burgoon2016nonverbal,schneider2017autism}. If adaptation occurs, it can 
be transient, subtle, multifold, and variant over time, which makes the quantitative 
analysis of the 
adaption process a challenging task. A possible approach to deal with this problem 
would  be to present the nonverbal adaptation process in a form of time series of 
features and then perform a  cause-effect analysis on the obtained time series. Among 
the many known causality inference methods, \gls{GC} \cite{GRANGER1980329} is the most 
widely used one. \gls{GC}  states that causes both precede and help 
predict their effects. It  has been applied in a variety of scientific fields, 
such as finding causes for stock price changes in economics 
\cite{granger2000bivariate}, attribution of climate change in climate informatics 
\cite{zhang2011causality}, and analysing neural interactions in neuroscience 
\cite{ding2006granger}. With respect to nonverbal human behavior, \gls{GC} was for 
example used to model dominance effects in social interactions 
\cite{kalimeri2011automatic}, focusing on vocal and kinesic cues. 
Novel developments in computer vision and social signal processing yielded accurate, 
open-source, real-time toolboxes to easily extract facial expressions from images and 
videos. These easily accessible visual cues facilitate video and image analysis, not 
only in terms of segmentation and classification but can also be used to identify  
social cause-effect relationships. Surprisingly,  the capabilities of computer vision and 
social signal processing have rarely been combined.
In our work, we will exploit computer vision capabilities for a quantitative 
verification of hypotheses on cause-effect relations in real data by investigating 
time series of facial expressions via facial muscle activation or \gls{AUS} 
\cite{ekman2002facial}. The 
real data was obtained from an experimental setup in which dyadic dialogs between 
participants were recorded with one participant being instructed to behave in a 
particular way. Given the experimental setup, we expected the instructed person to 
cause the uninstructed person with regards to certain facial expressions. As the 
facial adaption process in dyadic dialogs is a complex process, specific novel methods 
had to be used.

The novel contributions of our study can be summarized as follows.
\begin{enumerate}
\item Exploiting computer vision methods, we provide a comprehensive concept for 
analysing 
the direction of influence in dyadic dialogs starting with raw video material.

\item Interaction implies mutual influence and causality. Causal inference concepts, such 
as \gls{GC}, have been rarely used to identify the direction of influence in nonverbal 
emotional communication. To the best of our knowledge no other work has used a Granger 
causality model to identify the direction of influence regarding facial expressions in 
dyadic dialogs.

\item Facial \gls{AUS} go along with emotional experience. However, in constructed 
situations distinct strong emotions might not be visible at all and a single \gls{AU} 
does not contain enough information for inferring emotions. We present applicable 
features when strong distinct facial emotions are seldom  visible. By using \gls{AUS} we 
derive facial expressions in upper and lower face regions from the six basic 
emotions \cite{ekman1992argument}. 

 \item We propose a method for the selection of the relevant time intervals where 
\gls{GC} should be applied, and show based on synthetic as well as real data, the 
superiority of the proposed method in detecting cause-effect relations when compared 
to applying GC on the full  time series.
\end{enumerate}

\section{\uppercase{Related Work}}
\noindent The topic of finding causal structures in nonverbal communication data is 
addressed by Kalimeri et al. \cite{kalimeri2012modeling}. In their paper, they used 
\gls{GC} for modeling the effects that dominant people might induce on the nonverbal 
behavior (speech energy and body motion) of other people. Besides audio 
cues, motion vectors and residual coding bit rate features from skin colored regions 
were extracted. In two systems, one for body movement and another one for speaking 
activity, with four time series each, a small \gls{GC} based causal network was 
used to identify the participants with high or low causal influence. Unlike our approach, 
the authors did not use facial expressions and do not identify relevant intervals in a 
previous step, but use the entire time series instead.

A popular approach for the latter strategy is to find similar segments, for example 
emotions, arousal or (dis)agreement, in videos. The literature holds several approaches 
that pose complex classification tasks. Kaliouby and Robinson \cite{el2005real} provided 
the first classification system for agreement and disagreement as well as other 
mental states based on nonverbal cues only. They used head motion and facial action 
units together with a dynamic Bayesian Network for classification. Also, a survey on 
cues, 
databases, and tools related to the detection of spontaneous agreement and disagreement 
was done by Bousmalis et al. \cite{bousmalis2013towards}.  Despite their ingenious 
methods, these approaches do not investigate cause-effect relations in the social 
interaction situation. Sheerman-Chase et al. \cite{sheerman2009feature} used visual cues 
to distinguish between states such as thinking, understanding, agreeing, and questioning 
to recognize agreement.

Matsuyama et al. \cite{matsuyama2016socially} developed a socially-aware robot 
assistant responding to visual and vocal cues. For visual features, the robot 
extracted facial cues (based on OpenFace \cite{baltrusaitis2018openface}) such as 
landmarks, head pose, gaze, and facial action units. Conversational strategies that 
build, 
maintain, or destroy budding relationships were classified. Moreover, rapport was 
estimated by temporal association rule learning. The researchers' approach investigates 
building a social relationship between a human and a robot; however this study does not 
deal with a time variant direction of cause-effect relation.

\section{\uppercase{Methodology}}
Multiple steps are  necessary to get from raw video material of dyadic dialogs to 
measuring the adaption process of interacting partners. In the following subsections, 
we will explain all these steps starting with the experimental setup and video 
recording in subsection \ref{sec:ExperimentalSetup}. In subsection 
\ref{sec:FeatureExtraction}, we introduce the nonverbal communication features, 
which represent the raw time series, extracted from the video material.
In subsection \ref{sec:GrangerCausality}, we introduce our \gls{GC} 
model and in subsection \ref{sec:InervalSelection} the algorithm used for 
transferring raw time series to time series consisting of relevant information 
only. Finally, in subsection \ref{ref:ModelCauseEffectRelations} we 
combine all previous steps to elucidate our entire approach.

\subsection{Experimental Setup}
\label{sec:ExperimentalSetup}
\begin{figure}[!h]
\centering
 \includegraphics[width=0.5\textwidth]{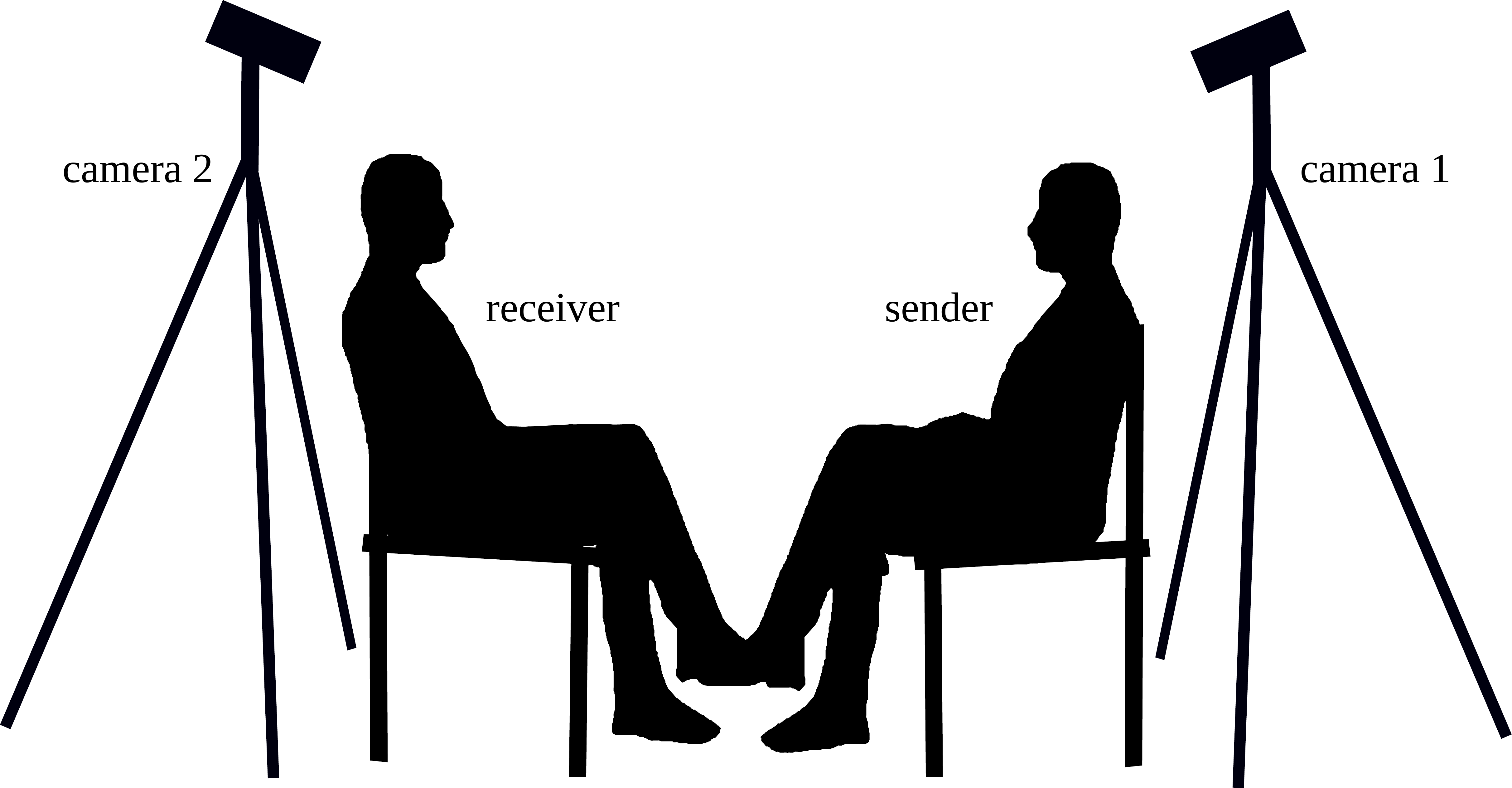}\caption{Experimental 
setup with camera positions showing sender and receiver sitting opposite to each other.}
\label{fig:ExperimentalSetup}
\end{figure}

\noindent We created an experimental setup (Figure \ref{fig:ExperimentalSetup}) in which 
two participants sat opposite to each other while talking about their personal weaknesses 
for about four minutes at a time. 

\noindent In total, they were asked to do this three times, either in 
circumstances of a respectful, contemptuous, or objective situation. One participant 
was in the assigned role of a \gls{R}, the other in the assigned role of the \gls{S}. 
As only \gls{S} had the active experimental interaction attitude task (i.e., to 
behave either respectfully, objectively, or contemptuously), we expected \gls{S} to 
influence \gls{R}  in relevant facial expressions. In all three experimental conditions 
each participant kept their initially assigned role of acting as a sender or receiver and 
the experimental conditions were conducted in a counterbalanced order. Further, R was 
asked to start the conversation with a personal weakness and both participants were asked 
to talk about at least one weakness per condition. The psychological research question 
was, whether and how S and R influence each other under the different attitude situations 
and how each interaction partner evaluates each dyadic interaction in terms of 
self-reported positive affect, liking, authenticity, engagement, and experienced 
self-other overlap. In order to avoid flirtatious situations, that may overwrite the 
instructed condition, interaction partners were always from the same sex. In total, 13 
pairs of participants (4 males; 9 females) were analysed in terms of their nonverbal 
behavior. All were students that participated for a small incentive (i.e., some 
chocolate) 
or course credit. All participants gave written informed consent. The study was 
conducted in accordance with the Declaration of Helsinki and approved by the Ethics 
Committee of the Friedrich Schiller University of Jena.

To capture nonverbal facial behavior, we positioned two frontal perspective cameras 
(Figure \ref{fig:ExperimentalSetup}), recording at 25 frames per second. Camera positions 
and lighting conditions were optimized during a test session before the study started. 
This ensured high video quality in terms of a plain frontal view of the faces and 
two-sided illumination. Motion blur rarely occurred, but could not be prevented entirely,
especially in cases of faster movements like head turns. Except for the experimental 
condition label no other information (e.g., expression annotation per frame) were 
available for image analysis. The entire dataset consists of 13 pairs, 
three conditions each pair and about 4 minutes of video per condition, thus about 300 
minutes of video material or 470.000 images.

\subsection{Facial Expressive Feature Extraction}
\label{sec:FeatureExtraction}
\noindent According to Ekman and Rosenberg \cite{ekman1997face}, facial expressions are 
the 
most important nonverbal signal when it comes to human interaction. The \gls{FACS} 
was developed by Ekman and Friesen \cite{ekman1978facial,ekman2002facial}. It 
specifies facial \gls{AUS}, based on facial muscle activation. Examples of \gls{AUS} 
are the \textit{inner brow raiser}, the \textit{nose wrinkler}, or the \textit{lip 
corner puller}. Any facial expression is a combination of facial muscles being 
activated, and thus, can be described by a combination of \gls{AUS}. Hence, the six 
basic emotions (\textit{anger}, \textit{fear}, \textit{sadness}, \textit{disgust}, 
\textit{surprise}, and \textit{happiness}) can also be 
represented via \gls{AUS}. Langner \cite{langner2010presentation} show that when for 
example \gls{AU} 6 (\textit{cheek raiser}), 12 (\textit{lip corner puller}), and 25 
(\textit{lips part}) are activated \textit{happiness} is visible.

In general, emotions are visual nonverbal communication cues transferable to time 
series. Regarding our real experimental data, this approach is reasonable for positive 
emotions like \textit{happiness}, which is frequently visible throughout the dyadic 
interactions. Yet, it is not applicable for negative associated emotions such as 
\textit{anger, disgust, fear,} or \textit{sadness}, as these emotions were only slightly 
visible in the dyadic interactions which may be due to the constructed experimental 
situation (Table \ref{tab:emotionCounts}).

\begin{table}[!ht]
\centering
\caption{Percentage of frames where emotions were visible throughout experiment.}

\begin{tabular}{|c|c|}
  \hline
  Emotion & Detection (in \%) \\
  \hline
  Happiness & 12.25 \\
  Surprise & 0.94 \\
  Anger & 0.13 \\
  Disgust & 3.72 \\
  Fear & 0.05 \\
  Sadness & 1.40 \\
  \hline

\end{tabular}
\label{tab:emotionCounts}
\end{table}

The approach of using stand-alone \gls{AUS} has two disadvantages. 
First, we cannot deduce emotional expressions from single \gls{AUS}. Second, 
lower \gls{AUS} are frequently activated while talking, and thus, are less suitable for 
analysis when it comes to emotional relations in dyadic interactions.

Wegrzyn et al. \cite{wegrzyn2017mapping} studied the relevance of facial areas for 
emotion classification and found differences in the importance of the eye and mouth 
regions. Facial \gls{AUS} can be divided into upper and lower \gls{AUS} 
\cite{cohn2007observer}. Upper \gls{AUS} belong to the upper half of the face and 
cover the eye region, whereas \gls{AUS} in the lower face half cover the mouth 
region. Hence, we decided to split emotions into upper and lower emotions, 
according to the affiliation of \gls{AUS} to upper and lower face regions. For 
example, instead of using \textit{sadness} as a combination of \gls{AU}1, \gls{AU}4, 
\gls{AU}15 and \gls{AU}17 we used \textit{sadness upper} (\gls{AU}1 and \gls{AU}4) and 
\textit{sadness lower} (\gls{AU}15 and \gls{AU}17). We only kept \textit{happiness} as a 
combination of both, upper and lower \gls{AUS}, as it was very frequently detected. All 
other emotions were split according to their \gls{AUS} belonging to upper or lower facial 
half (Table \ref{tab:expressionAndAUs}). This procedure ensured, that also subtle facial 
expressions were detectable and identified as an emotion.

\begin{table}[!h]
\caption{Expressions and corresponding \gls{AUS}.}\label{tab:expressionAndAUs} 
\centering
\begin{tabular}{|c|c|}
\hline
Expression            & Active Action Units   \\
\hline
Happiness            & 6, 12        \\
Surprise upper & 1, 2, 5     \\
Surprise lower & 26           \\
Disgust lower  & 9, 10, 25    \\
Fear upper     & 1, 2, 4, 5   \\
Fear lower     & 20, 25       \\
Sadness upper  & 1, 4         \\
Sadness lower  & 15, 17       \\
Anger upper    & 4, 5, 7      \\
*Anger lower   & 17, 23, 24  \\
\hline
\end{tabular}
\newline
\newline
\centering
*As \gls{AU}24 is not detected by OpenFace we excluded \textit{anger lower} 
from further analysis.
\end{table}

\begin{table}[!ht]
\centering
\caption{Percentage of emotions in upper and lower face parts visible throughout 
experiment.}
\label{tab:emotionCountsUpperLower}
\begin{tabular}{|c|c|}
  \hline
  Emotion & Detection (in \%) \\
  \hline
  \textit{ Anger lower } & 9.42 \\
  \textit{ Anger upper } & 1.42 \\
  \hline
  \textit{ Disgust lower } & 3.72 \\
  \hline
  \textit{ Fear lower } & 4.35 \\
  \textit{ Fear upper } & 1.12 \\
  \hline
  \textit{ Happy lower } & 16.12 \\
  \textit{ Happy upper } & 26.51 \\
  \hline
  \textit{ Sadness lower } & 8.74 \\
  \textit{ Sadness upper } & 7.25 \\
  \hline
  \textit{ Surprise lower } & 26.41 \\
  \textit{ Surprise upper } & 2.22 \\

  \hline

\end{tabular}
\end{table}

\begin{figure}[!th]
\centering
\includegraphics[width=0.45\textwidth]{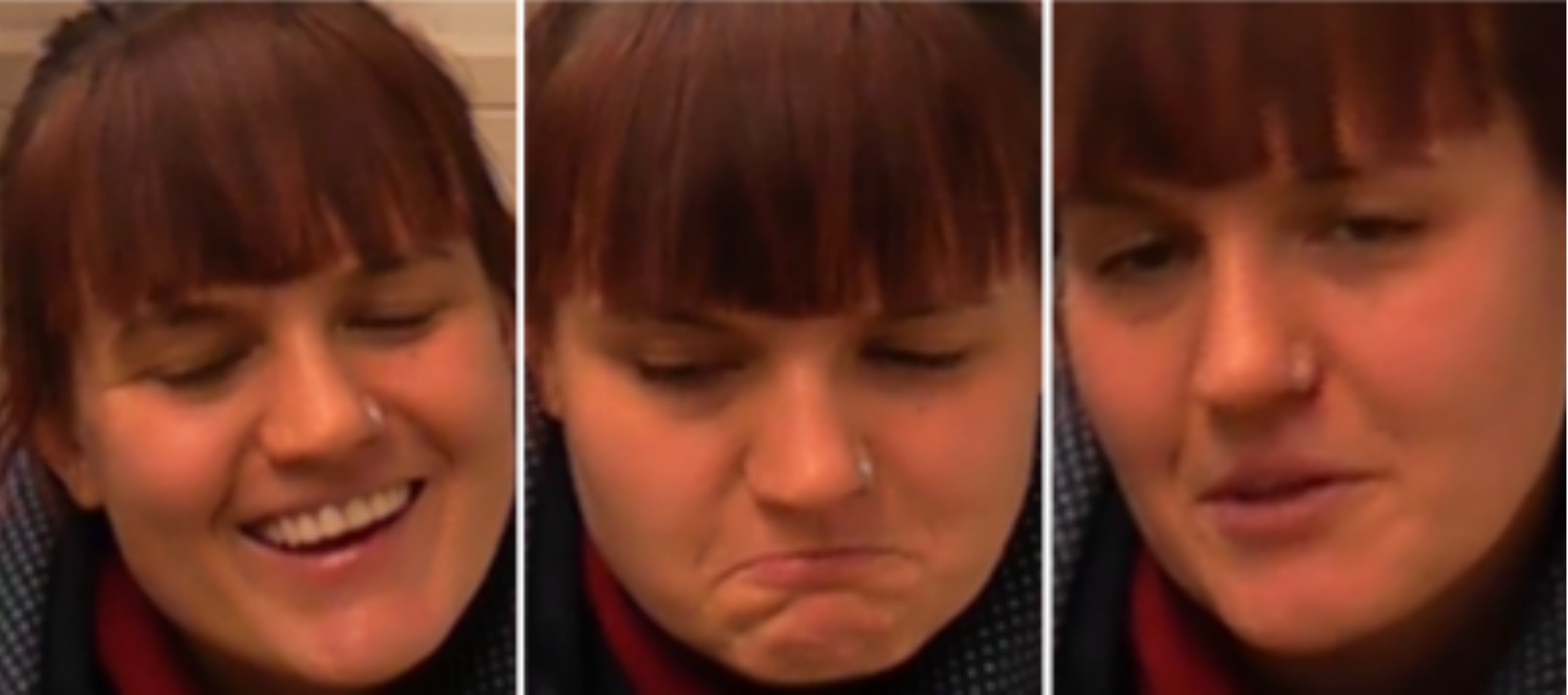}\caption{
From left to right participant with \gls{AU} 6 and 12 (\textit{happiness}), \gls{AU} 
15 
and 17 (\textit{sadness lower}), and \gls{AU} 1 and 4 (\textit{sadness upper}) being 
activated.}
\label{fig:AUSActivation}
\end{figure}

In Table \ref{tab:emotionCountsUpperLower} the detection percentage of upper and lower 
expressions 
is illustrated. After splitting, \textit{anger lower}, \textit{sadness lower}, 
\textit{sadness upper}, and \textit{surprise lower} emotions were detected in over 7 
\% of 
the video material on average. Figure \ref{fig:AUSActivation} illustrates which upper and 
lower expressions are detected based on the \gls{AU} activation for \textit{happiness}, 
\textit{sadness upper}, and \textit{sadness lower}.

\begin{figure}[!th]

\centering
\begin{subfigure}

\includegraphics[width=0.45\textwidth,height=3cm, 
keepaspectratio]{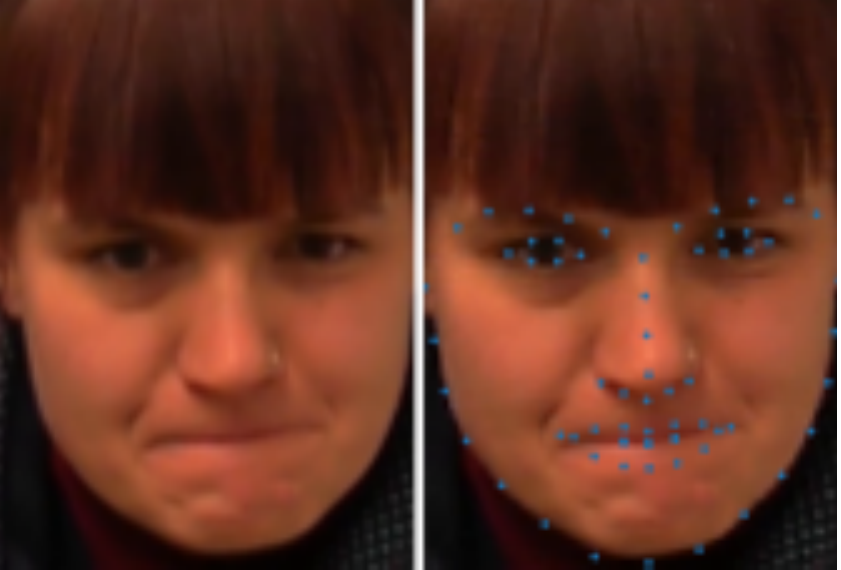}
\end{subfigure}

\begin{subfigure}

\includegraphics[width=0.45\textwidth]{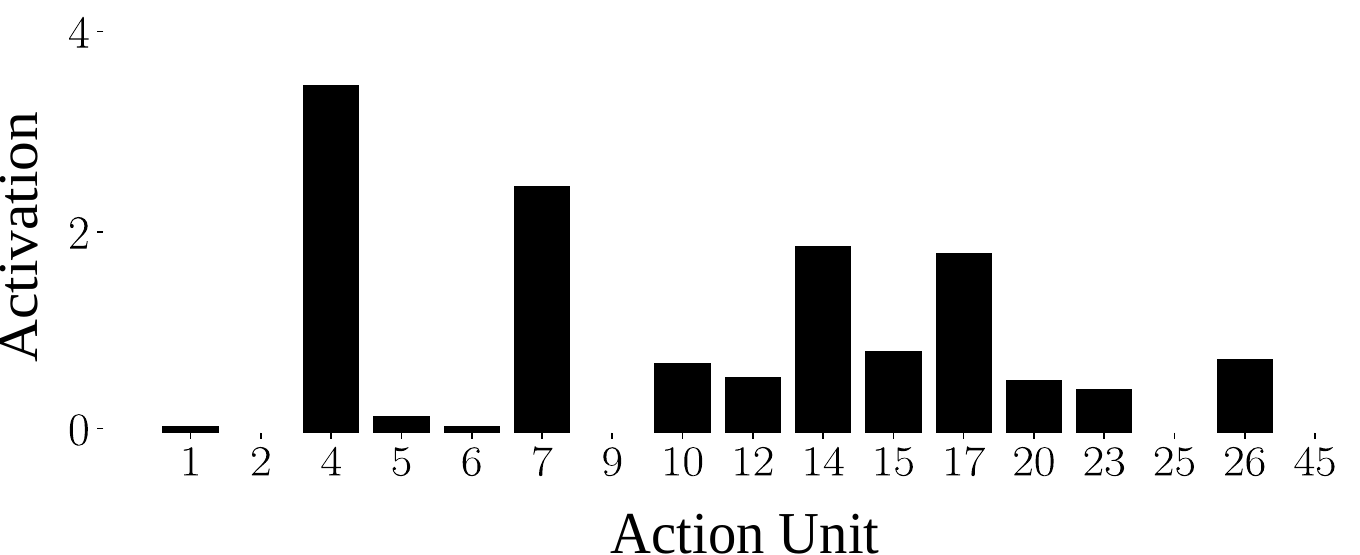}\label{
fig:AUDetectionOnly}
\end{subfigure}
\caption{Facial expression with landmarks and \gls{AUS} detected by 
OpenFace. Strong activation of \gls{AU}4 (\textit{brow lowerer}), 7 (\textit{lid 
tightener}), 14 (\textit{dimpler}), and 17 (\textit{chin 
raiser})}\label{fig:AUDetection}
\end{figure}

For feature extraction, we used OpenFace 
\cite{baltrusaitis2018openface,baltruvsaitis2015cross} which is a state of the art, 
open-source tool for 
landmark detection; it estimates \gls{AUS} based on landmark positions. OpenFace is 
capable of extracting 17 different \gls{AUS} (1, 2, 4, 5, 6, 7, 9, 10, 12, 14, 15, 
17, 20, 23, 25, 26, 45) with an intensity scaled from 0 to 5. Figure 
\ref{fig:AUDetection} illustrates the detection of landmarks and \gls{AUS} 
for an example image.

s

\subsection{Granger Causality}
\label{sec:GrangerCausality}
\noindent For the purpose of finding cause-effect relations, Granger's concept of 
causality was used. \gls{GC} is based on the axiom of temporal precedence, meaning that 
the past and present may cause the future, but the future cannot cause the past 
\cite{GRANGER1980329}. 

Let $\mathbf{x}_t = (x_1, x_2 \dots x_z)_t$ and $\mathbf{y}_t = (y_1, x_2 \dots 
y_z)_t$ 
be real-valued $z$-dimensional (column) vectors of \gls{AUS} at time point $t$, $t=1 
\dots T$, and let $\bar{x}_t = \frac{1}{z} \sum_{i=1}^{z} (x_i)_t$ and $\bar{y}_t = 
\frac{1}{z} \sum_{i=1}^{z} (y_i)_t$ be the average of  $\mathbf{x}_t$ and 
$\mathbf{y}_t$ 
at time point $t$. This results in two time series $\mathbf{X}_t = \bar{x}_1, \dots 
\bar{x}_T$ and $\mathbf{Y}_t = \bar{y}_1, \dots \bar{y}_T$ consisting of averaged values 
of \gls{AUS}. For building the (averaged) \gls{GC} model we require $\mathbf{X}_t$ and 
$\mathbf{Y}_t$ to be stationary.

The prediction of values of $\mathbf{X}$ and $\mathbf{Y}$ at time $t$ is based on 
previous values from $\mathbf{X}_k$ and $\mathbf{Y}_k$, $k<t$
$$\mathbf{X}_{t} = \sum_{j=1}^m a_{j}\mathbf{X}_{t-j} + \sum_{j=1}^m 
b_{j}\mathbf{Y}_{t-j} + \varepsilon_t 
\quad (1)$$
$$\mathbf{Y}_{t} = \sum_{j=1}^m c_{j}\mathbf{X}_{t-j} + \sum_{j=1}^m 
d_{j}\mathbf{Y}_{t-j} + \vartheta_t \quad 
(2)$$

with $\varepsilon_t$ and $\vartheta_t$ being two independent noise processes. For each 
expression of each participant in each condition we estimated the best model order 
$m$ using the \gls{BIC}. For statistical significance, an F-Test with a level of 
significance of $p=0.05$ was used. When testing 
for \gls{GC} three different cases regarding the direction of influence can occur 
\cite{schulze2004granger}:
\begin{enumerate}
 \item If $a_k = 0$ for $k = 1 \dots m$ and $\exists b_k \neq 0$ for $1 \leq k 
\leq m$ then $\mathbf{Y}$ Granger causes $\mathbf{X}$.
\item If $d_k = 0$ for $k = 1 \dots m$ and $\exists c_k \neq 0$ for $1 \leq k 
\leq m$ then $\mathbf{X}$ Granger causes $\mathbf{Y}$.
\item If for both $\exists b_k \neq 0$ for $1 \leq k \leq m$ and $\exists c_k \neq 
0$ for $1 \leq k \leq m$ holds a bidirectional (feedback) relation exists.
\end{enumerate}

\noindent If none of the above cases holds, $\mathbf{X}$ and $\mathbf{Y}$ are 
not Granger causing each other. In our real data, we expected that, if present, 
pairs that do not Granger cause each other are rare.  

\subsection{Relevant Interval Selection}
\label{sec:InervalSelection}
\noindent Considering the experimental setup, we had to expect multiple temporal 
scenes, further referred to as subintervals, in which the participants influenced each 
other. The time spans where causality is visible, might range from half a second to 
half a minute, occur several times, and can be interrupted by irrelevant scenes 
(e.g., one participant talking while the other participant is listening) that differ in 
the 
length of time. As outlined above, the direction of influence in a subinterval can either 
be bidirectional, or unidirectional driven by either \gls{S} or \gls{R}. This implies 
that three unwanted effects can occur, if the full time span is analysed: first, temporal 
relations are not found at all; second, bidirectional relations mask temporal 
unidirectional relations and; third, an unidirectional relation from $\mathbf{X}$ to 
$\mathbf{Y}$ masks temporal bidirectional influence or unidirectional influence from 
$\mathbf{Y}$ to $\mathbf{X}$. Li et al. \cite{li2017discovery} give an example where  
temporal \gls{GC} is not being detected, when the full time span is used for model 
fitting.

Our central idea is to apply \gls{GC} only to time series obtained by concatenating 
highly coherent (e.g., in terms of Pearson correlation) subintervals of raw time 
series. Instead of using a brute force algorithm, we suggest using a bottom-up 
approach for finding the longest set of maximal, non-overlapping, correlated intervals in 
time series as proposed by Atluri et al. \cite{atluri2014discovering}. The authors 
applied their approach to fMRI data where they achieved good results for clustering 
coherent working brain regions.

Let $\mathbf{X}_t$ and $\mathbf{Y}_t$ be two time series of length $N$. An interval 
is called \textit{correlated interval} for a threshold $\beta$, when all its 
subintervals up to a lower interval length $\alpha$ are correlated as well. An 
interval $I_{(a,b)}$ from $a$ to $b$ is called maximal, when $I_{(a,b)}$ is a 
\textit{correlated interval}, but $I_{(a-1, b)}$ and $I_{(a,b+1)}$ are not. 
And two intervals $I_{(a,b)}$ and $I_{(c,d)}$ are called non-overlapping, when $I_{(a,b)} 
\cap I_{(c,d)} = \emptyset$. From all intervals fulfilling these conditions the longest 
set (total length of intervals) is computed. 

In a multivariate case (e.g., multiple \gls{AUS} defining an expression), we propose to 
compute the longest set for each pair of corresponding variables and then use the 
intersection of intervals over all variables of the system, as selected 
relevant intervals. For further analysis, for each variable of the system the selected 
relevant intervals can be concatenated, resulting in multiple time series each composed 
of relevant information only. In the following we refer to the set of selected intervals 
between two time series $\mathbf{X}$ and $\mathbf{Y}$ as $AW_{XY}$.

\subsection{Modeling Cause-Effect Relations}
\label{ref:ModelCauseEffectRelations}
\noindent The two major challenges in the analysis of the cause-effect relations in 
dyadic dialogs, that make the application of conventional methods difficult were:
\begin{enumerate}
 \item Due to the constructed situations, strong distinct emotions, computed by using 
traditional \gls{AU} combinations, were rarely visible.
\item Time variant and situation-dependent communication, resulting in a high 
variety and volatility of time spans in which nonverbal cause-effect behavior 
between interacting partners is visible.
\end{enumerate}

\noindent  To tackle these difficulties, we use the combination of specific facial 
expressions and the relevant interval selection approch. The final selection of 
relevant intervals and the following analysis of causality for two systems of facial 
action units $\mathbf{x}_1 \dots \mathbf{x}_T$ and $\mathbf{y}_1 \dots \mathbf{y}_T$
consists of the following steps:
\begin{enumerate}
 \item Calculate selected relevant intervals $AW_{x_{1t}y_{1t}}, AW_{x_{2t}y_{2t}}, \dots 
, AW_{x_{zt}y_{zt}}$ pairwise between corresponding system parameters.
 \item Calculate the intersection $AW_{\mathbf{x}\mathbf{y}}$ of all sets of 
selected intervals 
$AW_{x_{1t}y_{1t}} \cap AW_{x_{2t}y_{2t}} \cap \dots \cap AW_{x_{zt}y_{zt}}$.
\item Concatenate selected intervals for each variable of $\mathbf{x}_t$ and 
$\mathbf{y}_t$
\item Compute \gls{GC} on concatenation.
\end{enumerate}
%facial expression time series X and Y
\noindent Before applying the relevant interval selection approach to our nonverbal 
communication data, we identified upper and lower facial expressions that changed 
significantly between the three experimental conditions. For that we calculated each 
participant's \textit{average face}, which is the average 
\gls{AU} activation over the three conditions and used it as a lower threshold for the 
activation of an expression. That means, that we considered an expression as 
\textit{visible}, when all of its associated \gls{AUS} were greater than 0.5 standard 
deviations of the conspecific \gls{AUS} in the \textit{average face}. The number of 
activations per 
expression was counted per person and experimental condition, and normalized by video 
length and maximum count of the expression of each person. A Wilcoxon signed-rank 
test revealed, that the participants showed significantly more \textit{happiness} in the 
respectful condition than in the contempt condition ($p = .034$, $s = 92.0$). Further, we 
found both, more \textit{sadness lower} ($p = .034$, $s = 92.0$) and \textit{sadness 
upper} ($p = .023, s = 86.0$) expressions in the contempt condition compared to the 
objective condition, when using a Benjamini-Hochberg p-value correction 
\cite{benjamini1995controlling} with a false discovery rate of $Q = .3$ and 
individual p-values of $\alpha = .5$.

As a next step, we applied the relevant interval selection approach, for computing 
selected intervals, pairwise to all of the identified \gls{AUS}, with a minimum interval 
length of 75 and a threshold of 0.8 for Pearson 
correlation. Based on known average human reaction time (ca. 200 ms or 6 frames 
\cite{jain2015comparative}), we shifted one time series by 0, 4, 8, and 12 
frames both, back and forth in time, and computed relevant intervals. The grid 
selected for 
shifting does cover quicker and slower reactions of participants, while being 
computationally performant. Afterwards, we computed the longest set of the list of 
relevant intervals obtained from the different shifts. Before computing \gls{GC}, we 
median filtered the selected intervals with a filter length of 51 (2 seconds). Finally, 
we calculated the average \gls{GC} on the concatenation of the intervals in the set of 
selected intervals of the smoothed (Gaussian blur with $\sigma^2 = 1$, window size $20$) 
standardized time 
series. The results were counted according to the possible outcomes of the \gls{GC} test 
in \ref{sec:GrangerCausality}, as either unidirectional caused by \gls{S}, unidirectional 
caused by \gls{R}, bidirectional, or no causality.

\section{\uppercase{Experimental Results and Discussion}}
\label{sec:results}

\noindent \textbf{Evaluation on synthetic data.} The following constructed example 
illustrates, how our idea contributes to a better detection of coherent subintervals 
in time series. Initially, we generated two time series of length $N=6000$, so that 
$\mathbf{X}_t, \mathbf{Y}_t \sim {\mathcal{N}}(0,1)$, and $\mathbf{X}_t$, $\mathbf{Y}_t$ 
are independent. We then smoothed (Gaussian blur with $\sigma^2 = 1$, window size 
$10$) $\mathbf{X}$ and $\mathbf{Y}$. After that, multiple intervals of random length 
$l_s$, $l_s \sim {\mathcal {U}}(50,200)$ were synchronized and $\mathbf{Y}$ shifted 
by four samples back in time. A synchronized interval is followed by an 
unsynchronized interval of length $l_u$, $l_u \sim {\mathcal {U}}(100,600)$. In the 
last step, we added Gaussian noise $\epsilon$ to $Y$, $\epsilon \sim 
\mathcal{N}(0,0.02)$.

\begin{figure}[!ht]
\centering
  
\includegraphics[width=0.45\textwidth]{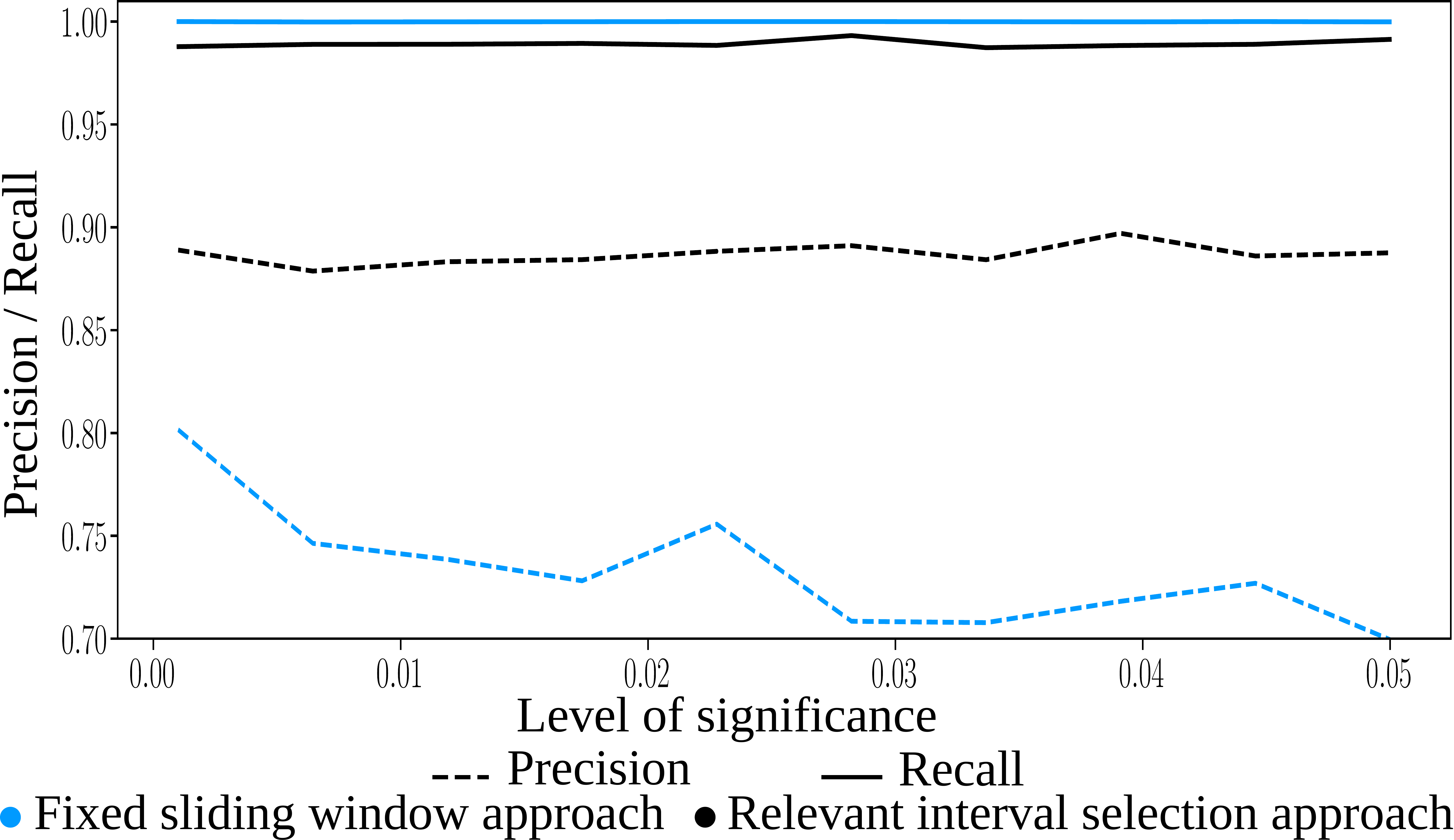}
  \caption{Precision and recall for relevant interval selection and fixed size sliding 
window approaches for $\mathbf{Y}$ Granger causing $\mathbf{X}$.}
  \label{fig:CompareLamiFixedXcy}
\end{figure}

\begin{figure}[!ht]
\centering
 
\includegraphics[width=0.45\textwidth]{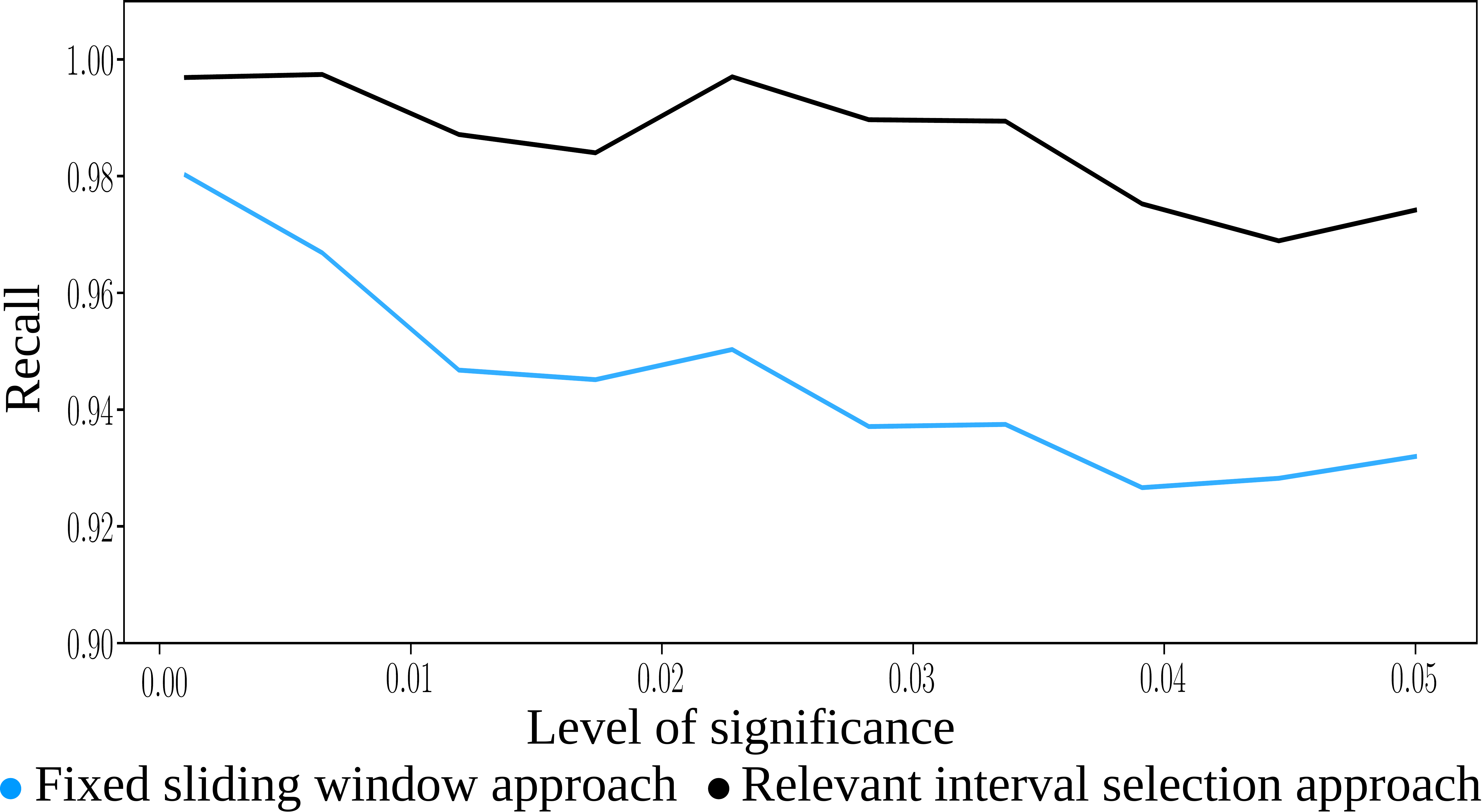}
 \caption{Recall for relevant interval selection and fixed size sliding window 
approaches for $\mathbf{X}$ Granger causing $\mathbf{Y}$.}
\label{fig:CompareLamiFixedYcX}
\end{figure}

We expected the following approaches to detect all synchronized intervals, and 
identify 
the cause-effect relation on each interval in the manner that $\mathbf{Y}$ Granger 
causing $\mathbf{X}$, and no intervals for $\mathbf{X}$ Granger causing 
$\mathbf{Y}$, at 
different levels of significance $\alpha$. We compare the following two approaches:
\begin{enumerate}
 \item \textit{Fixed size sliding window approach:} For the fixed size sliding window 
approach we used window 
size $\gamma = 50$ and step size $\nu = 2$. Since multiple tests are performed, a 
Bonferroni corrected p-value $p_f = \frac{\alpha\nu}{2(n-\gamma)}$ was used for 
detecting \gls{GC}.
 \item \textit{Relevant interval selection approach:} We set the minimum windows size  
to 50, the correlation threshold to $0.9$, and used a two-sided time shift of 4. 
The Bonferroni corrected p-value 
$p_{aw} = \frac{\alpha}{2\left\vert{AW_{XY}}\right\vert}$ was selected according to 
the number of intervals $\left\vert{AW_{XY}}\right\vert$ detected by the relevant 
interval selection approach.
\end{enumerate}
For $\mathbf{Y}$ Granger causing $\mathbf{X}$, we evaluated precision and recall with 
the 
synchronized intervals as ground truth. For $\mathbf{X}$ Granger causing $\mathbf{Y}$, 
the 
ground truth is the full time series, and thus, only recall needs to be evaluated. Figure 
\ref{fig:CompareLamiFixedXcy} shows the evaluation for $\mathbf{Y}$ Granger causing 
$\mathbf{X}$. Both approaches show a very good performance in detecting all relevant 
intervals (recall). Yet, the relevant interval selection approach detects less irrelevant 
intervals (precision) among all levels of significance. Figure 
\ref{fig:CompareLamiFixedYcX} shows that both, relevant interval selection and fixed size 
sliding window approach, show a very high recall for $\mathbf{X}$ Granger causing 
$\mathbf{Y}$ among all levels of significance, but the relevant interval selection 
approach is slightly superior.
\newline

\begin{figure*}[!ht]
\centering
\includegraphics[width=\textwidth]
{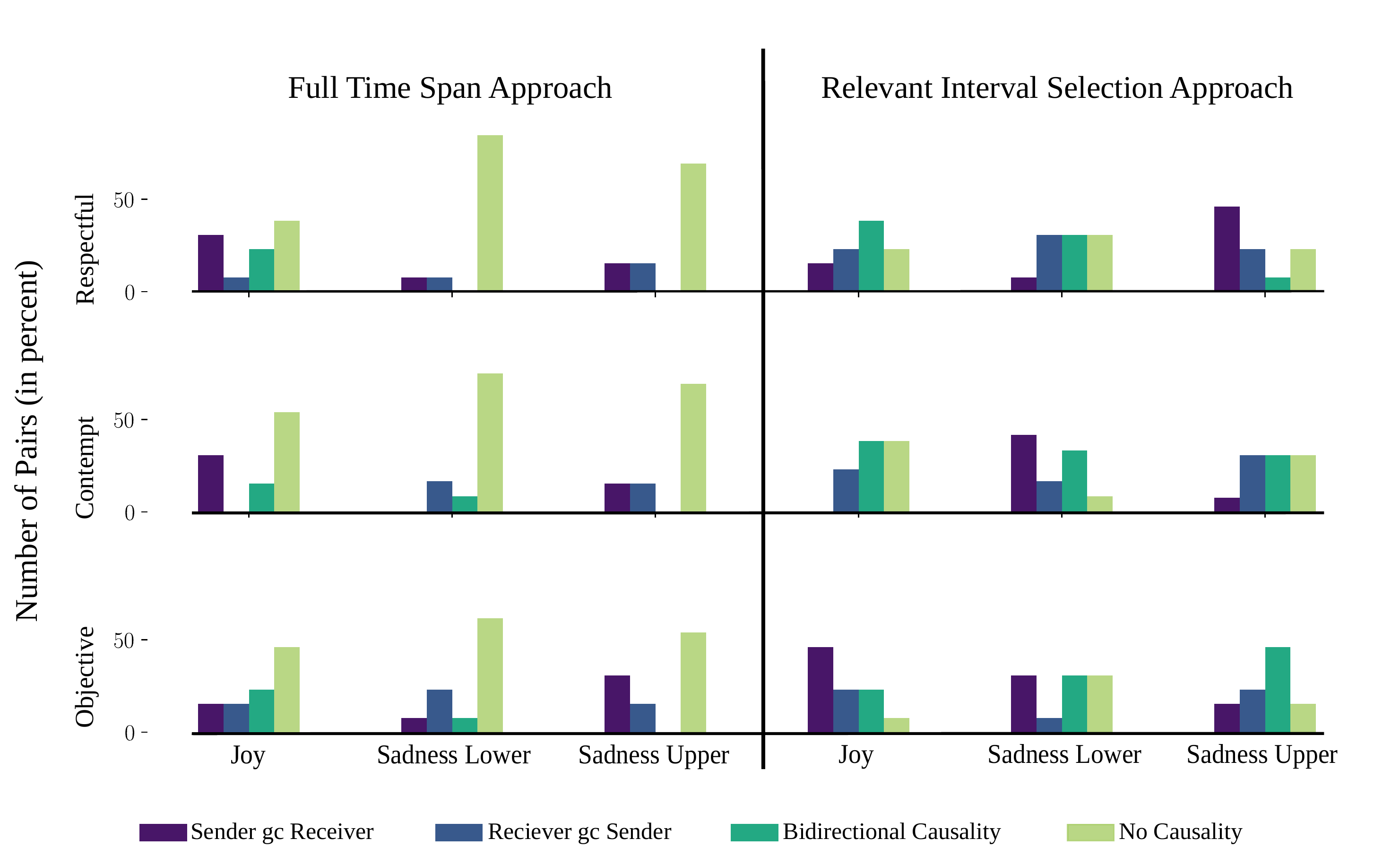}
  \caption{Relevant interval selection approach compared to full time span approach for 
different facial expressions under the three experimental conditions for 
distinguishing the direction of influence.}
  \label{fig:AdapFullResult}
\end{figure*}

\noindent \textbf{Evaluation on nonverbal communication data.} In Figure 
\ref{fig:AdapFullResult}, our relevant interval selection approach is 
compared to the full 
time span approach. The figure shows the percentage of pairs for which the \gls{GC} 
test, with $p = 0.05$, showed a specific direction of influence, under the three 
experimental conditions, for each of the identified expressions (\textit{sadness lower}, 
\textit{sadness upper}, \textit{happiness}). Especially for \textit{sadness lower} and 
\textit{sadness upper} expressions, the full time span approach does not find causality 
between \gls{S} and \gls{R} for over 50\% of the pairs. With our relevant interval 
selection approach, less pairs show \textit{no causality}, but instead uni- or 
bidirectional causation. Especially in the \textit{sadness lower} condition, we were able 
to detect that the direction of influence was more often driven by \gls{S} or 
bidirectional, and rarely driven by \gls{R}. The full time span approach does not 
expose this information at all. In the contempt condition \gls{S} is not supposed to 
show positive expressions.

\section{\uppercase{Conclusions}}
\label{sec:conclusion}
In this paper, we employed \gls{GC} together with a relevant interval selection approach 
on synthetic and nonverbal communication data obtained from an experimental 
setup. Based 
on the results of Wegrzyn et al. \cite{wegrzyn2017mapping} we designed our own emotional 
facial features, capable of capturing emotions even when 
strong distinct emotions are not visible. Our facial expressions are composed of facial 
action units, which can be detected by real-time, state of the art computer vision tools. 
We proposed an intelligent interval selection approach for filtering relevant information 
in dyadic dialogs. Subsequently, we were able to apply our \gls{GC} model to the 
concatenation of relevant intervals and compute the direction of influence.

We applied our approach to real data. On the one hand, this emphasised the 
superiority of the relevant interval selection approach compared to a full time span 
approach and on 
the other hand, this revealed that in contemptuous dyadic dialogs \textit{happiness} was 
more often caused by \gls{R} whereas \textit{sadness lower} was more often initiated by 
\gls{S} - a pattern of results which was theoretically to be expected. In general many 
bidirectional influences were found.

As no standardized mapping from \gls{AUS} to emotions exists, we proposed using a system 
of upper and lower facial expressions. For further research, we suggest using a 
learning system, capable of classifying upper and lower emotions based on all \gls{AUS}.

In our approach we used correlation which is a linear similarity measure. For nonlinear 
dependence, Pearson correlation can be easily replaced by appropriate distance measures 
such as mutual information. The \gls{GC} model must be changed accordingly.

Overall, we identify our contribution as an important step towards interdisciplinary, 
with computer vision potentials, psychological observations, and theoretical 
knowledge of causality methods being combined and extended to gain interesting 
insights into emotional social interaction.

\bibliographystyle{apalike}
{\small
\bibliography{bibliography}}

\vfill
\end{document}